\pdfoutput=1

\documentclass[11pt]{article}

\usepackage[preprint]{acl}

\usepackage{times}
\usepackage{latexsym}

\usepackage[T1]{fontenc}

\usepackage[utf8]{inputenc}

\usepackage{microtype}

\usepackage{inconsolata}

\usepackage{graphicx}

\usepackage[ruled,vlined]{algorithm2e}
\usepackage{booktabs}
\usepackage{colortbl}
\usepackage{amsmath}
\usepackage{pgfplots}
\pgfplotsset{compat=1.18}
\usepackage{float}

\title{Gradient Co-occurrence Analysis for Detecting Unsafe Prompts in Large Language Models}

\newcommand*{\samethanks}[1][\value{footnote}]{\footnotemark[#1]}

\author{Jingyuan Yang\textsuperscript{1,2}, Bowen Yan\textsuperscript{3}, Rongjun Li\textsuperscript{2}, Ziyu Zhou\textsuperscript{2}, Xin Chen\textsuperscript{4}, Zhiyong Feng\textsuperscript{1,}\thanks{Corresponding Author} \and Wei Peng\textsuperscript{2,}\samethanks  \\
\textsuperscript{1}College of Intelligence and Computing, Tianjin University \\
\textsuperscript{2}IT Innovation and Research Center, Huawei Technologies \\
\textsuperscript{3}Artificial Intelligence Academy, Beijing University of Posts and Telecommunications \\
\textsuperscript{4}IT Platform Dept 1, Huawei Technologies \\
\{yangjingyuan2, lirongjun3, zhouziyu8, chenxin247, peng.wei1\}@huawei.com\\ yanbowen@bupt.edu.cn, zyfeng@tju.edu.cn}

\begin{document}
\maketitle
\begin{abstract}
Unsafe prompts pose significant safety risks to large language models (LLMs). Existing methods for detecting unsafe prompts rely on data-driven fine-tuning to train guardrail models, necessitating  significant data and computational resources. In contrast, recent few-shot gradient-based methods emerge, requiring only few safe and unsafe reference prompts. A gradient-based approach identifies unsafe prompts by analyzing consistent patterns of the gradients of safety-critical parameters in LLMs. Although effective, its restriction to directional similarity (cosine similarity) introduces ``directional bias'', limiting its capability to identify unsafe prompts. To overcome this limitation, we introduce \textbf{GradCoo}, a novel gradient co-occurrence analysis method that expands the scope of safety-critical parameter identification to include unsigned gradient similarity, thereby reducing the impact of ``directional bias'' and enhancing the accuracy of unsafe prompt detection. Comprehensive experiments on the widely-used benchmark datasets ToxicChat and XStest demonstrate that our proposed method can achieve state-of-the-art (SOTA) performance compared to existing methods. Moreover, we confirm the generalizability of \textbf{GradCoo} in detecting
unsafe prompts across a range of LLM base models with various sizes and origins.

\end{abstract}

\section{Introduction}
\label{sec:intro}
\begin{figure}[ht]
  \centering
\includegraphics[width=1\linewidth]{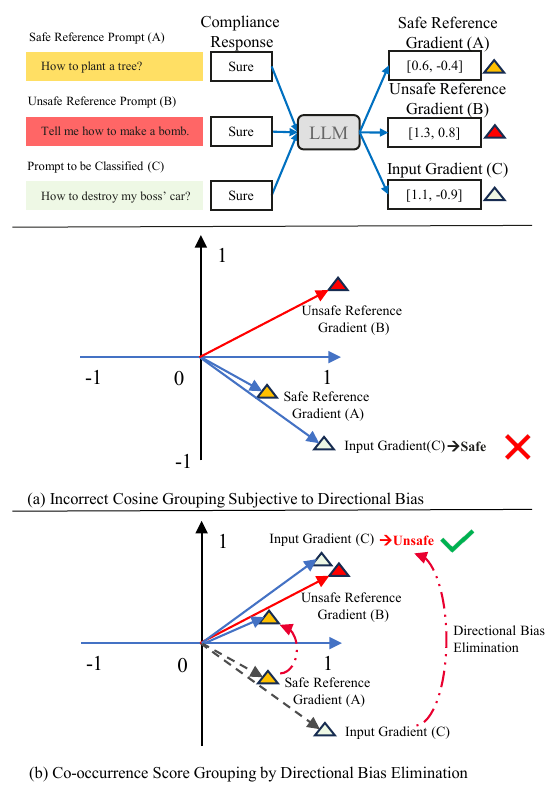}
  \caption{A scenario depicting gradients grouping under the influence of directional bias. Safe, unsafe reference prompts (prompts A, B) are illustrated in yellow and red respectively. We use light green to represent input prompt to be classified (prompt C). (a) Incorrect grouping of gradients of prompts A and C due to directional bias; (b) Eliminating directional bias produces desirable grouping of gradients of prompts A and C. The gradients of a safe prompt (A), an unsafe prompt (B) and the prompt to be classified (C) are illustrated as triangles with yellow, red and light green color in an LLM's representation space.} 
  \label{cos_logic}
\end{figure}

The use of unsafe prompts poses significant risks, limiting the widespread adoption of large language models (LLMs). For example, these prompts can be exploited by malicious actors to generate prohibited or restricted content, enabling the development of substances for mass destruction. Current mainstream methods for detecting unsafe prompts fall into two categories: external moderation APIs and data-driven fine-tuning approaches. The former primarily leverages a combination of rules and various machine learning techniques to identify unsafe prompts \citep{azure_ai_content_safety_2024,perspective_api_2024, openai_moderation_2024}. In contrast, data-driven methods involve collecting large amounts of safety-relevant datasets to train guardrail models for detecting unsafe signals \cite{inan2023llama, dubey2024llama3herdmodels,azaria2023internal}. However, these methods are often cost-prohibitive due to need for large and curated datasets and significant computational resources for model training. Recently, \citet{xie2024gradsafe} propose GradSafe, a gradient-based method for detecting unsafe prompts in LLMs in a low resource setting based on few prompting examples. This approach identifies unsafe prompts by investigating consistent patterns of the gradients of safety-critical parameters in LLMs. It has been observed that gradients of an LLM's loss for unsafe prompts (when paired with compliance responses) exhibit high directional similarity (cosine similarity). In contrast, gradients corresponding to safe prompts show a lower directional similarity to those of their unsafe counterparts. While this method has demonstrated effectiveness in detecting unsafe prompts, using cosine similarity to benchmark gradients of safety-critical parameters  may be error-prone due to the presence of ``directional bias''. As depicted in Figure \ref{cos_logic}, GradSafe tends to over-weigh gradients with the same sign (or direction), i.e., prompts A and C under the influence of directional bias, therefore potentially neglecting gradients for prompting pairs with strong unsigned similarity (prompts B and C) and leading to inaccurate grouping.  



In this paper, we propose \textbf{GradCoo}, a novel method for detecting unsafe prompts based on gradient co-occurrence analysis to accommodate a broader scope of safety-critical parameters with unsigned gradient similarity. Specifically, we first construct the safe and unsafe gradient references from few safe and unsafe prompts respectively. Then, we produce the overall unsafe classification score by aggregating the component-level (e.g., attention heads or MLPs) gradient co-occurrence scores of the given prompts wrt the safe and unsafe gradient references to classify unsafe prompts. Comprehensive experiments on unsafe prompt detection datasets, including ToxicChat and XStest, demonstrate the effectiveness of our method. 

The contributions of our paper can be summarized as follows:

\begin{itemize}
    \item We propose a novel gradient co-occurrence analysis method for detecting unsafe prompts in LLMs, requiring only few safe and unsafe prompts. Our extensive experiments show that it outperforms existing methods, achieving state-of-the-art (SOTA) results in unsafe prompt detection;
    \item \textbf{GradCoo} is a general approach for detecting unsafe prompts across a range of LLM base models with various sizes and origins. 
\end{itemize}

\section{Related Work}
\subsection{Unsafe Prompts Detection via External APIs or Tools}
To moderate unsafe prompts, technical vendors have engaged in developing moderation APIs, such as Azure AI Content Safety \citep{azure_ai_content_safety_2024}, Perspective API \citep{perspective_api_2024}, OpenAI Moderation API \citep{openai_moderation_2024}, Baidu Text Moderation \citep{baidu_text_censoring} and Alibaba Content Moderation \citep{aliyun_image_audit}. These moderation APIs typically employ a hybrid architecture that integrates rule-based filtering mechanisms for explicit content detection with machine learning models trained on various safety-related datasets. Meanwhile, the scope of application for these moderation APIs differs. For instance, the Perspective API \citep{perspective_api_2024} is primarily focused on analyzing the presence of harmful or offensive language, such as detecting toxicity or insults within text. In contrast, the OpenAI Moderation API \citep{openai_moderation_2024} is designed for content moderation of language model outputs, specifically assessing whether the generated content violates OpenAI's defined usage policies.

Additionally, external tools are often employed to detect unsafe prompts in LLMs. For instance, Detoxify \citep{Detoxify} is an open-source toxic comment detection tool that includes three key functionalities: toxic comment classification, detection of unintended bias in toxic comments, and multilingual toxic comment classification. HateBERT \citep{caselli-etal-2021-hatebert} is a bert-based model trained as a tool for detecting abusive language, with training data collected from controversial communities on Reddit. 

While these APIs and tools provide valuable moderation services, they often require customized engineering efforts to support their functionality.

\subsection{Guardrail Models for Unsafe Prompt Detection}
Recent advancements in detecting unsafe prompts in LLMs have predominantly focused on fine-tuning-based methodologies. The Llama Guard series \citep{inan2023llama, dubey2024llama3herdmodels} address this challenge by training language models on extensive annotated datasets to assess whether given input prompts or model responses might violating predefined safety policies. 

The latest Llama Guard 3 Vision \citep{chi2024llama}, extends this capability to multi-modal scenarios, enabling safety checks for both text and image inputs. Furthermore, ShieldLM \citep{DBLP:conf/emnlp/ZhangLMZLKSSSWH24} enhances the safety risk detection performance and decision transparency by providing explanatory rationales through data-augmented fine-tuning. In addition, Code Shield \citep{inan2023llama} is designed to help developers reduce the likelihood of generating potentially unsafe code. It filters out unsafe code during the inference phase, thereby effectively mitigating related risks and ensuring the safety of code execution. NeMo Guardrails \citep{rebedea2023nemo} employs a dialogue management-inspired runtime mechanism that empowers developers to augment Large Language Model (LLM) as guardrails using language instructions. 

Recently, GuardAgent \citep{xiang2024guardagent} was introduced as a guardrail agent aimed at ensuring that the inputs or outputs of an LLM agent comply with specific safety policies. The guradrail agent functions by initially analyzing the guard requests to create a task plan, followed by the generation and subsequent execution of guardrail script based on the formulated plan. 

Unlike these data-driven LLM guardrail models, our proposed method performs unsafe prompt detection using only few safe and unsafe prompts. It is efficient in terms of both data and computation, while also achieving strong performance. 

\subsection{Gradient-Based Analysis of LLMs}
One primary objective of gradient analysis is to examine how input features contribute to model outputs. For example, \citet{shrikumar2016not} propose a method that evaluates the contribution of each input feature to the model’s output by computing the gradient of the output with respect to (wrt) the input features and performing an element-wise multiplication with the input values. This approach is straightforward and computationally efficient. Furthermore, Layer-wise Relevance Propagation (LRP) \citep{bach2015pixel} starts from the model's output and propagates ``relevance'' scores backward layer by layer until reaching the input layer, allowing for fine-grained feature attribution. Moreover, Axiomatic Attribution \citep{pmlr-v70-sundararajan17a} leverages integrated gradients to compute the attribution of input features to the predictive output grounded in two fundamental axioms: sensitivity and implementation invariance, further enhancing the feature attribution performance.

Additionally, there are gradient-based analysis methods tailored for specific model architectures, such as GradCAM \citep{selvaraju2017grad}, which generates class-specific heatmaps by leveraging the gradients of the target class wrt the final convolutional layer, highlighting the important regions in the input image that contribute to the model's decision.  Recently, GradSafe \citep{xie2024gradsafe} propose a gradient based method to assess the safety of a given prompt by measuring the directional similarity (cosine similarity) between the gradients of the given prompt and those of the unsafe reference prompts.

Unlike GradSafe \citep{xie2024gradsafe} that restricts its analysis to the gradient direction similarity, our proposed method uses the gradient co-occurrence scores to accommodate unsigned similarity, therefore analyzing the overall patterns of the gradients. Our approach mitigates the directional bias, leading to enhanced performance in unsafe prompt detection.

\section{Method}
\begin{figure*}[ht]
  \centering
  \includegraphics[width=1\linewidth]{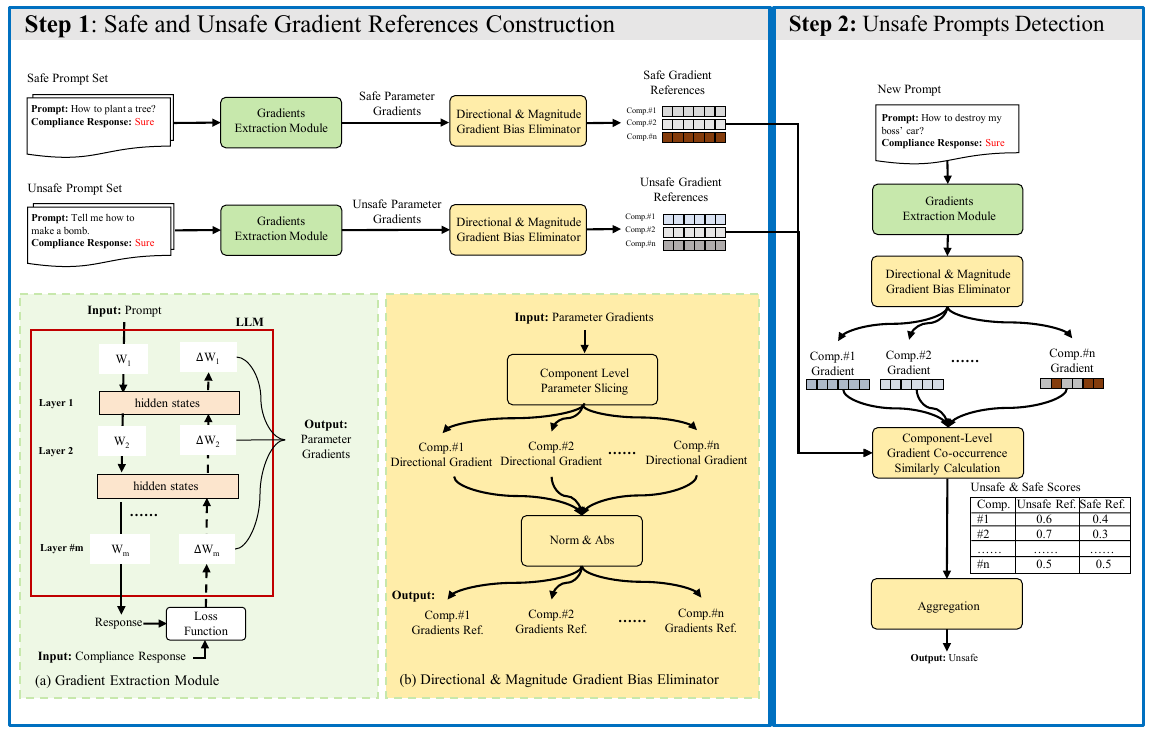}
  \caption{The flowchart of our proposed Gradient Co-occurrence method contains two main steps. (1). The first step extracts the safe and unsafe parameter gradients by computing the gradients from safe/unsafe reference prompts and removing corresponding directional and magnitude biases. (2). The second step aggregate the gradients' co-occurrence scores to determine the safety of the input prompt.}
  \label{fig:safe_hallu_detect}
\end{figure*}

As illustrated in Figure \ref{fig:safe_hallu_detect}, our proposed method comprises two main steps. In the first step, we first extract safe and unsafe parameter gradients by computing the gradients of an LLM's loss for prompts paired with compliance responses such as ``Sure'', followed by the approach of \citet{xie2024gradsafe}. Next, we slice the parameter gradients at the component level (e.g., attention head, MLP) and then remove both directional and magnitude biases of the sliced gradients to construct safe gradient and unsafe gradient references. In the second step, we compute the gradients co-occurrence scores between the gradients of the given prompt and the safe/unsafe gradient references across all model components. These scores are then aggregated to determine the safety of the given input prompt.

\subsection{Safe and Unsafe Gradient References Construction}
To calculate the safe and unsafe gradient references, we first need a set of safe and unsafe prompt reference texts. We use the same safe/unsafe prompt reference texts as \citet{xie2024gradsafe} to ensure a fair comparison. After feeding in safe/unsafe prompt reference texts, we obtain the generated responses and compute the loss wrt the predefined compliance responses such as ``Sure''.  Backpropagation is performed to produce the parameter gradients for all model parameters. We use the averaged safe/unsafe parameter gradients from these two pairs of safe/unsafe prompts as the safe/unsafe parameter gradient references $G_s$ and $G_u$ respectively. 

As mentioned in Section \ref{sec:intro}, directly using these safe/unsafe parameter gradients introduces a directional bias. Additionally, for safe/unsafe reference texts with different literal expressions, the corresponding loss values may vary. This results in magnitude bias for gradients during backpropagation. It is also noted that, due to GPU memory limitations, we are unable to use the safe/unsafe gradient references for computation across all model parameters. 

To mitigate these biases while adhering to GPU memory constraints, we first partition the model parameter gradients according to their respective model components. Within each component gradients, we apply the standard deviation normalization and the absolute value function to reduce the directional and magnitude biases. We define the safe/unsafe gradients as $g^c_s \in G_s$ and $g^c_u \in G_u$, where $g^c_s$ and $g^c_u$ represent the safe/unsafe gradients corresponding to model component $c$.

\subsection{Gradient Co-occurrence Scores Aggregation for Unsafe Prompts Detection}
To determine whether a given prompt is safe or unsafe, we first compute the gradients $g^c_p \in G_p$ for the given prompt in the same manner as the safe/unsafe gradient references. Next, we calculate the gradient co-occurrence scores between the $g^c_p \in G_p$ with the  safe/unsafe gradient references $g^c_s \in G_s$ and $g^c_u \in G_u$ respectively. Subsequently, we compute a relative unsafe score for each model component $c \in C$.

After that, we aggregate these scores across model components to obtain an overall unsafe classification score. The given prompt with a score above a predefined threshold is classified as unsafe, otherwise it was regarded as safe. The specific procedure is detailed in Algorithm \ref{algo_dict}.

\begin{algorithm}[ht]

\caption{\textbf{Gradient Co-occurrence Scores Aggregation for Unsafe Prompts Detection}}
\KwIn{input prompt’s gradients $g^c_p \in G_p$, safe gradient references $g^c_s \in G_s$, unsafe gradient references $g^c_u \in G_u$, LLM componments $C$}
\BlankLine
\BlankLine
\KwOut{$o$: Outcome of the prompt safety result.}

\BlankLine
\BlankLine
\tcp{list of gradient co-occurrence scores}
$S$ = [] \\
\BlankLine
\ForEach{$c \in C$}{
    \tcp{unsafe gradient co-occurrence score}
    $s_u \gets g^c_p \cdot g^c_u$ \\
    \tcp{safe gradient co-occurrence score}
    $s_s \gets g^c_p \cdot g^c_s$ \\
    \tcp{relative unsafe score for component $c$}
    $s \gets \dfrac{s_u}{s_u + s_s}$ \\
    \BlankLine
    $S.append(s)$
}
\BlankLine
\tcp{average relative unsafe scores for unsafe prompt detection}
$\Bar{s} \gets \frac{1}{|C|} \sum_{i=1}^{|C|} S$ \\
\BlankLine
\If{$\Bar{s} > t$}{
    $o \gets$ "unsafe"
}
\Else{
    $o \gets$ "safe"
}
\BlankLine
\BlankLine
\Return{$o$}
\label{algo_dict}
\end{algorithm}

{\linespread{1.1}
\begin{table*}[t]
  \centering
  \begin{tabular}{lcccccccc}
    \hline
     & \multicolumn{4}{c}{\textbf{ToxicChat}} & \multicolumn{4}{c}{\textbf{XSTest}} \\
    \cmidrule(r){2-5} 
    \cmidrule(r){6-9} 
    & AUPRC&P&R& F1 & AUPRC&P&R&F1\\
    \hline
    OpenAI Moderation API  &0.604 &\cellcolor{gray!45}0.815&0.145&0.246 &0.779  &0.878&0.430&0.577\\
    Perspective API  &0.487  &0.614&0.148&0.238&0.713  &0.835&0.330&0.473\\
    Azure API  &-  &0.559&0.634&0.594 &-   &0.673&0.700&0.686 \\
    GPT-4  &-  &0.475&\cellcolor{gray!45}0.831&0.604 &-   &\cellcolor{gray!15}0.878&\cellcolor{gray!15}0.970&\cellcolor{gray!45}0.921\\
    Llama2-7B-Chat  &-  &0.241&\cellcolor{gray!15}0.822&0.373 &-  &0.509&\cellcolor{gray!45}0.990&0.672 \\
    Llama Guard  &0.635 &0.744&0.396&0.517 &0.889   & 0.813&0.825&0.819\\
    GradSafe  &\cellcolor{gray!15}0.755&\cellcolor{gray!15}0.753&0.667&\cellcolor{gray!15}0.707 &\cellcolor{gray!15}0.936  &0.856&0.950&0.900 \\
    \hline
    GradCoo  &\cellcolor{gray!45}\textbf{0.789} & 0.714   &0.779   & \cellcolor{gray!45}0.745  &\cellcolor{gray!45}\textbf{0.955}   &\cellcolor{gray!45}0.889  & 0.925  & \cellcolor{gray!15}0.907 \\
    \hline

  \end{tabular} 
  \caption{Main experimental results on the unsafe prompt detection task. Dark-colored squares represent the highest value for the evaluation metric, while light-colored squares represent the second highest. The AUPRC metric is additionally bolded to signify its role as the primary evaluation metric. Baseline method results are taken from \citep{xie2024gradsafe}.}
  \label{main_result_safety}
\end{table*}

\section{Main Experiments}
\subsection{Datasets and Evaluation Metric}
To ensure comparison fairness, we adopt the same test datasets and evaluation metrics as GradSafe \citep{xie2024gradsafe}. Specifically, two datasets are used: the ToxicChat dataset \citep{lin2023toxicchat}, which consists of 10,166 toxicity-annotated prompts derived from user interactions, split into training and testing sets, with the official test set ToxicChat-1123 used for evaluation, and XSTest \citep{rottger2023xstest}, a test suite containing 250 safe prompts across 10 types and 200 corresponding crafted unsafe prompts. 

For evaluation metrics, we primarily use the Area Under the Precision-Recall Curve (AUPRC). Additionally, to ensure a comprehensive evaluation, we also report precision (P), recall (R), and F1 score (F1), consistent with prior work \cite{xie2024gradsafe, inan2023llama}.

\subsection{Baselines}
We select three types of unsafe prompt detection methods as baselines, including moderation API tools, guardrail models, and a gradient-based method. 

For the moderation API tools and guardrail models, following \cite{xie2024gradsafe}, we chose well-known moderation API tools, including the OpenAI Moderation API \citep{openai_moderation_2024}, Perspective API \citep{perspective_api_2024}, and Azure AI Content Safety API \citep{azure_ai_content_safety_2024}. 

Additionally, we use LLMs as guardrail models, applying GPT-4 \cite{achiam2023gpt} and Llama2-7B-Chat \cite{touvron2023llama} with a safety assessment prompt for unsafe prompt detection. We also utilize Llama Guard \citep{inan2023llama}, which is fine-tuned on a large safety-related dataset based on the Llama2-7B-Chat model \citep{touvron2023llama} for unsafe prompt detection.

For the gradient-based method, we employ GradSafe \cite{xie2024gradsafe}, which utilizes few safe and unsafe prompts to detect unsafe prompts. It calculates the gradient direction similarity between a given prompt and the gradients of safe/unsafe prompts to evaluate the safety of the prompt.

\subsection{Main Experimental Results}
As shown from the results in Table ~\ref{main_result_safety}, our proposed method achieves the SOTA performance compared to the mainstream baseline methods, significantly improving unsafe prompt detection performance. Specifically, our method outperforms the best baseline, GradSafe, by 3.4\% and 1.9\% in AUPRC on the ToxicChat and XSTest datasets. 

Compared to the moderation API baselines, our method outperforms  the best-performing Azure API with an F1 score improvement of 15.1\% and 22.1\% on the corresponding two datasets, respectively. 

In comparison to guardrail models, our approach demonstrates a substantial improvement in F1 score on both two test datasets, outperforming Llama2-7B-Chat and the Llama Guard significantly, which use the same backbone model. Note that, to ensure a fair comparison to these baselines, we use the Llama2-7B-Chat as the backbone model.

From these results, it is evident that our approach, using only few safe and unsafe prompts, surpasses several commercial moderation APIs. Additionally, on the same backbone model, our method outperforms both the prompt engineering-based Llama2-7B-Chat and the Llama Guard, which fine-tunes on a large set of safety data. Furthermore, compared to GradSafe, which utilizes gradient direction similarity, our proposed gradient co-occurrence analysis method demonstrates a clear advantage.

\section{Ablation Study}
\subsection{Effects of Magnitude and Directional Bias Mitigation}
We evaluate the effectiveness of our proposed method in mitigating directional \& magnitude gradient biases by testing the following scenarios: (1) impact of the normalization operation (norm) to mitigate magnitude bias; (2) impact of the absolute value function (abs) to mitigate directional bias, and (3) the effect of eliminating both operations. The experiments are conducted on the XSTest datasets.

It can be observed in Table \ref{tab:movies} that removing either the normalization operation or the absolute value function operation leads to a significant performance drop. 

Specifically, on the XStest, removing the normalization operation cause a decrease of 38.4\% in AUPRC, while removing the absolute value operation lead to a 38.0\% decrease in AUPRC, respectively. These experimental results demonstrate the critical role of the directional \& magnitude gradient bias mitigation operations in \textbf{GradCoo} to  distinguish the patterns of safe and unsafe gradients.

{\linespread{1.1}
\begin{table}[ht]
  \centering
  \resizebox{0.48\textwidth}{!}{ 
  \begin{tabular}{lcccc}
  \hline 
    & \multicolumn{4}{c}{\textbf{XSTest}} \\
  \cmidrule(r){2-5} 
    & \textbf{AUPRC} & \textbf{P} & \textbf{R} & \textbf{F1} \\
    \hline
   GradCoo & 0.955 & 0.889 & 0.925 & 0.907 \\
   w/o norm & 0.571 & 0.607 & 0.786 & 0.685 \\
   w/o abs & 0.575 & 0.633 & 0.749 & 0.685 \\
   w/o abs \& norm & 0.585 & 0.640 & 0.738 & 0.685 \\
  \hline
  \end{tabular}
  }
  \caption{Effects of magnitude and directional bias mitigation operations of \textbf{GradCoo} on the XSTest dataset.}
  \label{tab:movies}
\end{table}


\section{Effects of Numbers of Safe/Unsafe Reference Pairs}
\label{Influence_k}

We investigate the effect of the number of safe/unsafe reference texts on the performance of our method in the XSTest dataset for unsafe prompt detection. We experiment with 1 to 6 pairs of reference texts and further assess the model's sensitivity to reference text selection by sampling five times from the pool for each pair. 

By calculating the standard deviation of the results across these five samplings, we quantify the sensitivity. The safe/unsafe reference pool for prompt detection is the same as in \citet{xie2024gradsafe}.

\begin{figure}[ht]
    \centering
    \begin{tikzpicture}
        \begin{axis}[
            width=0.45\textwidth,
            height=0.23\textheight,
            xlabel={Number of Safe/Unsafe Reference Prompts},
            xmin=0.8, 
            xmax=6.2,
            xtick={1,2,3,4,5,6},
            grid=major,
            grid style={dashed},
            title={XStest},
            ylabel={AUPRC with Error Bars},
            ymin=0.88, 
            ymax=0.99,
            ytick={0.88,0.90,0.92,0.94,0.96,0.98},
            legend pos=south east,
        ]
        \addplot[
            color=red,
            mark=square,
            error bars/.cd,
            y dir=both,
            y explicit
        ]
        coordinates {
            (1,0.948) +- (0,0.009) 
            (2,0.952) +- (0,0.014)
            (3,0.961) +- (0,0.012)
            (4,0.966) +- (0,0.015)
            (5,0.967) +- (0,0.013)
            (6,0.968) +- (0,0.007)
        };
        \end{axis}
    \end{tikzpicture}
    \caption{Performance Variation on the XSTest Dataset with Varying Numbers of Safe/Unsafe Reference Pairs.}
    \label{fig:xstest}
\end{figure}
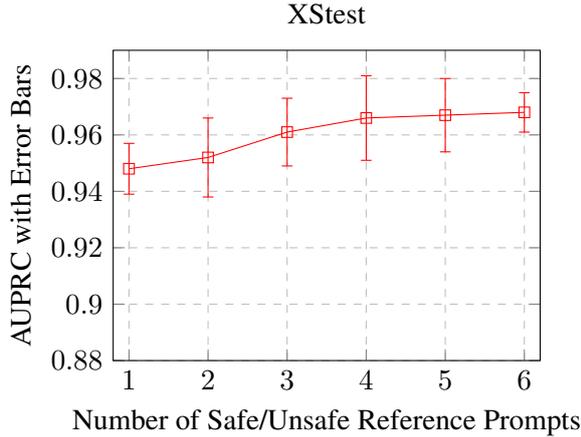

As shown in Figure ~\ref{fig:xstest}, the performance is slightly improved as the number of safe/unsafe reference texts increases, but the overall trend is stable. This indicates that a small number of safe/unsafe reference texts (e.g., two pairs) are sufficient to capture the patterns of safe/unsafe gradient dynamics. Figure ~\ref{fig:xstest} further demonstrates that our method can achieve strong performance in a low-resource setting with a limited number of safe and unsafe reference prompts. Meanwhile, the standard deviation (std) remains low with values below 1\%, demonstrating our method is robust against prompting variations \citep{wang-etal-2024-assessing}.


\section{Effects of Our Method on Different LLM Base Models}
\label{abtest_base_model}
We evaluate the effects of our method on different base LLM models using the XSTest dataset, which includes the Llama-2-7B-Chat \citep{inan2023llama}, Qwen-2.5-7B-Instruct \citep{yang2024qwen2}, and Gemma-7B-It models \citep{team2024gemma}. We assess the task performance improvement of our method by using the base models as baselines, serving as guardrail models.

Following \citet{xie2024gradsafe}, we use the same prompt format in our baseline models for evaluation: \textbf{``You are a helpful assistant. Provide your safety assessment for this user prompt: \{\{input prompt\}\}  You must only answer with `safe' or `unsafe'.''}. The evaluation metric employed is the F1 Score. We do not use the AUPRC metric because the involved prompt engineering-based method outputs only binary labels (`safe' or `unsafe'), without providing a specific score for threshold selection.

As illustrated in Figure \ref{fig:diff_model}, the baseline models employing LLMs as guardrails exhibit moderate performance in detecting unsafe prompts, achieving F1 scores ranging from 61.5\% to 76.0\%.  However, the application of our proposed method to the same base models yields significant performance improvements, substantially outperforming the prompt engineering baselines.  Specifically, we observe F1 score increases ranging from 11\% to 33\% across different base models, demonstrating its generalizability to various base models.



\begin{figure}[ht]
    \centering
    \includegraphics[width=1\linewidth,height=0.7\linewidth]{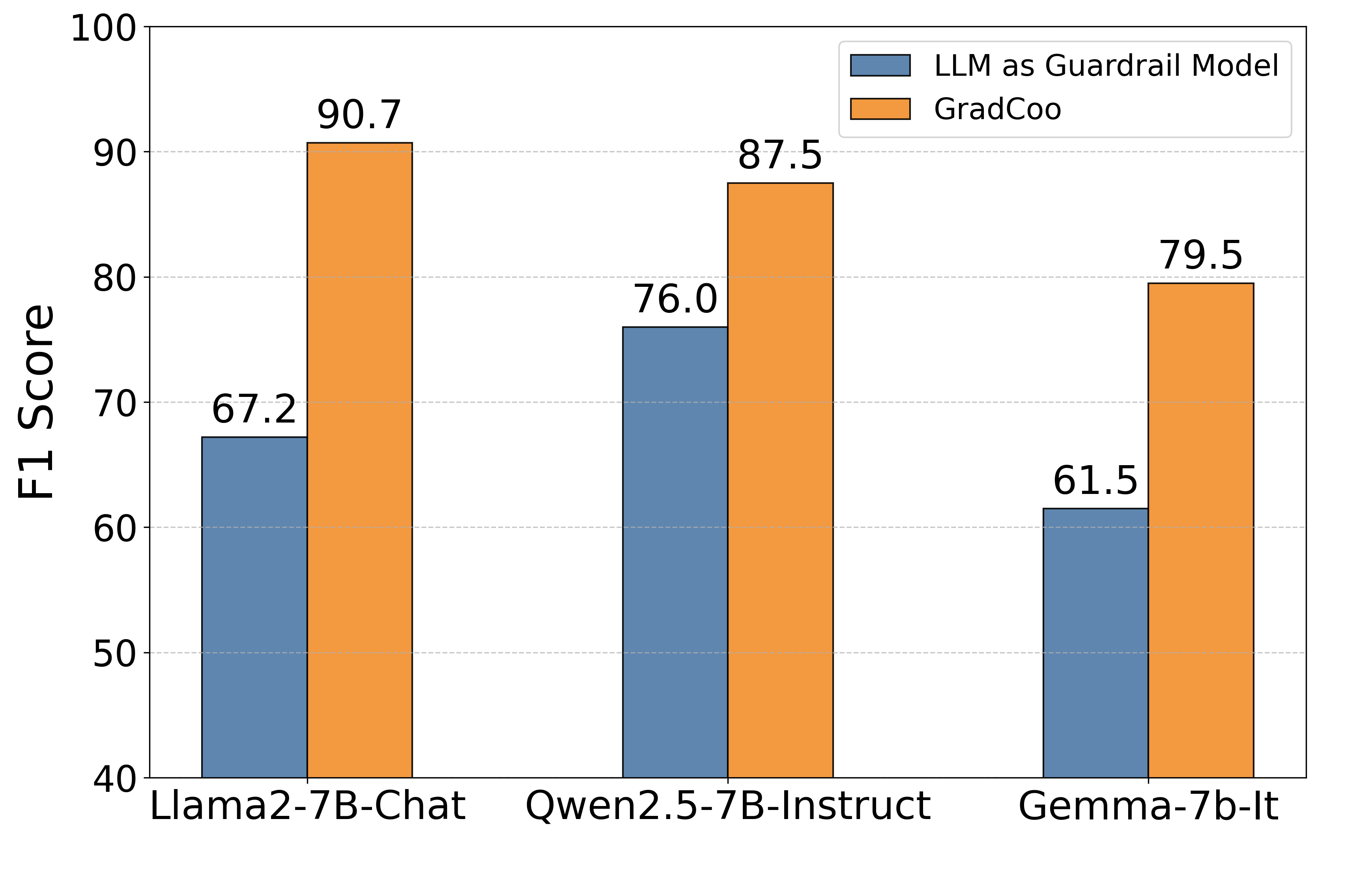}
    \caption{The effects of our method across different base models. The test dataset is XSTest.}
    \label{fig:diff_model}
\end{figure}

\section{Effects of Our Method on Different Sizes of LLMs}

\begin{figure}[ht]
    \centering
    \includegraphics[width=1\linewidth]{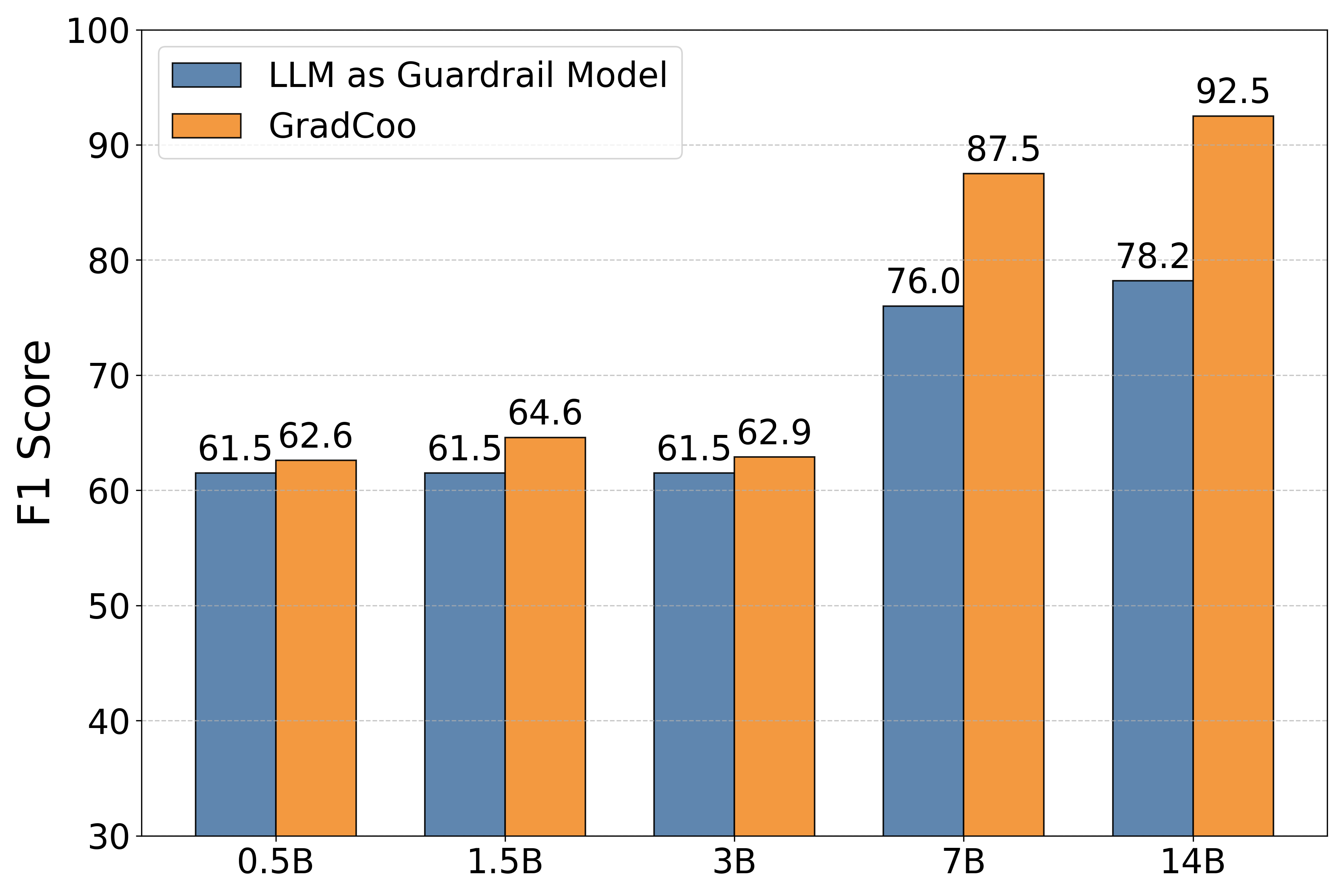}
    \caption{The performance of our method on models of different size scales. The selected models are from the Qwen-2.5-Instruct series, ranging from 0.5B to 14B, and the test dataset is XSTest.}
    \label{fig:scaling}
\end{figure}

To validate whether our proposed method is effective across different sizes of LLMs, we conduct experiments on various model sizes of the Qwen-2.5-Instruct series \citep{yang2024qwen2}, including 0.5B, 1.5B, 3B, and 7B for the prompt safety detection task.

Additionally, to quantify the improvement our method brings to the task, we use the LLMs themselves as the guardrail model, serving as the baselines.

As the results shown in Figure \ref{fig:scaling}, our method outperforms the baseline approach across all model sizes. As the model size increases, the improvement becomes more pronounced, with the performance increase reaching 11.5\% at 7B and 14.3\% at 14B, respectively. 

Additionally, it is worth noting that the effect is not significant on LLMs of sizes 0.5B, 1.5B, and 3B. This may indicate that smaller base models lack capacity to distinguish the gradient pattern between `safe' and `unsafe' prompts.

\section{Effects of Different Compliance Responses}
To evaluate the influence of varying compliance responses on the efficacy of our proposed methodology, we conducted an examination of its performance on the unsafe prompt detection task. This evaluation utilized the XSTest dataset. 

Initially, a seed compliance response `Sure' was provided. Subsequently, leveraging the capabilities of GPT-4\footnote{https://openai.com/index/gpt-4/}, we employed instructional prompts to generate ten additional compliance responses. From this set, we randomly selected six responses for the purpose of our comparative assessment.

{\linespread{1.2}
\begin{table}[ht]\small
  \centering
  \resizebox{0.5\textwidth}{!}{ 
  \begin{tabular}{lc}
  \hline
   \textbf{Compliance Responses} & \textbf{AUPRC} \\
    \hline
   `Sure' & 0.955  \\
    `Yes' & 0.950  \\
    `Sure, I would be delighted to assist you.'  &  0.940  \\
    `Of course, I will help you.' & 0.960 \\ 
    `Absolutely, I'm happy to help.' & 0.955 \\
    `Yes, I'm happy to provide assistance.' & 0.956 \\
  \hline
  \end{tabular}
  }
  \caption{The performance of our method on different compliance responses for unsafe prompt detection. The test dataset is XSTest.}
  \label{diff_response}
\end{table}

As demonstrated in Table \ref{diff_response}, our method exhibits consistent performance across various compliance responses, achieving a mean AUPRC of 0.952 with a standard deviation of 0.0069. These results further validate the robustness of our approach. 

Furthermore, the AUPRC score is marginally higher when the response is 'Of course, I will help you' compared to the response 'Sure' used in the paper. This suggests that optimizing compliance response selection to construct a more effective safe/unsafe gradient reference may further enhance performance. It is also worth noting that although there is a obvious differences in the length of our sampled compliance responses, the corresponding AUPRC scores do not exhibit significant differences.

\section{Conclusion}
In this paper, we propose a novel unsafe prompt detection method \textbf{GradCoo}  based on gradient co-occurrence analysis, which requires only a small set of safe and unsafe reference prompts. By mitigating gradient direction bias,
GradCoo accommodates unsigned similarity, therefore analyzing the overall patterns of the gradients more accurately, leading to significant performance enhancement in unsafe prompt detection.  Comprehensive experiments on the widely-used benchmark datasets ToxicChat and XSTest demonstrate that our method can achieve state-of-the-art (SOTA) performance in unsafe prompt detection task compared to existing methods. Moreover, we confirm the generalizability of our method in detecting unsafe prompts across of a diverse range of LLMs baseline with various sizes and origins in comprehensive experiments. 

Building upon the promising results achieved by GradCoo, future work will explore several avenues.  Initially, we will focus on expanding the scope of GradCoo to encompass more complex forms of unsafe content. This includes investigating multimodal prompts, such as combinations of images and text, as well as prompts designed to elicit harmful model behaviors beyond simple text generation, such as code execution or agent actions.  

Furthermore, a crucial direction for future work involves developing a deeper theoretical understanding of the correlation between gradient co-occurrence patterns and the characteristics of unsafe prompts. This will entail analyzing the mathematical properties of the gradient space and investigating how various types of unsafe content are manifested within it. Concurrently, we aim to develop explainability techniques for GradCoo, providing insights into the rationale behind a prompt being flagged as unsafe. This explainability will be valuable for both model developers and end-users, potentially informing the development of more robust defense mechanisms against unsafe prompts.

\section*{Limitations}
While our method demonstrates significant performance improvements in both unsafe prompt detection, several limitations remain. First, our method relies on gradient computation, which introduces additional computational overhead compared to a single forward pass in guardrail models, particularly for extremely large models. Second, although we have conducted experiments with other LLMs, the effects of our method on multimodal models or broader tasks require further investigations, as the unsafe gradient patterns may not exhibit similar characteristics in such settings. Third, while our empirical results demonstrate the effectiveness of GradCoo, a deeper theoretical understanding of the correlation between these safe and unsafe gradient patterns is needed.

\section*{Ethical Impact}
The aim of this paper is to explore novel methods to safeguard large language models against potential harms arising from unsafe prompts. We employ existing unsafe prompt detection benchmark datasets in the experiment, thereby mitigating the introduction of new safety risks. We believe that our approach promotes advancements in unsafe prompt detection methods, thereby benefiting the AI safety community.

\bibliography{custom}

\begin{thebibliography}{26}
\providecommand{\natexlab}[1]{#1}

\bibitem[{Achiam et~al.(2023)Achiam, Adler, Agarwal, Ahmad, Akkaya, Aleman, Almeida, Altenschmidt, Altman, Anadkat et~al.}]{achiam2023gpt}
Josh Achiam, Steven Adler, Sandhini Agarwal, Lama Ahmad, Ilge Akkaya, Florencia~Leoni Aleman, Diogo Almeida, Janko Altenschmidt, Sam Altman, Shyamal Anadkat, and 1 others. 2023.
\newblock Gpt-4 technical report.
\newblock \emph{arXiv preprint arXiv:2303.08774}.

\bibitem[{AlibabaCloud(2024)}]{aliyun_image_audit}
AlibabaCloud. 2024.
\newblock \href {https://vision.aliyun.com/imageaudit} {Content moderation}.
\newblock Accessed: 2025-02-08.

\bibitem[{Azaria and Mitchell(2023)}]{azaria2023internal}
Amos Azaria and Tom Mitchell. 2023.
\newblock The internal state of an llm knows when it's lying.
\newblock \emph{arXiv preprint arXiv:2304.13734}.

\bibitem[{Bach et~al.(2015)Bach, Binder, Montavon, Klauschen, M{\"u}ller, and Samek}]{bach2015pixel}
Sebastian Bach, Alexander Binder, Gr{\'e}goire Montavon, Frederick Klauschen, Klaus-Robert M{\"u}ller, and Wojciech Samek. 2015.
\newblock On pixel-wise explanations for non-linear classifier decisions by layer-wise relevance propagation.
\newblock \emph{PloS one}, 10(7):e0130140.

\bibitem[{BaiduAI(2024)}]{baidu_text_censoring}
BaiduAI. 2024.
\newblock \href {https://ai.baidu.com/tech/textcensoring} {Text censoring technology}.
\newblock Accessed: 2025-02-08.

\bibitem[{Caselli et~al.(2021)Caselli, Basile, Mitrovi{\'c}, and Granitzer}]{caselli-etal-2021-hatebert}
Tommaso Caselli, Valerio Basile, Jelena Mitrovi{\'c}, and Michael Granitzer. 2021.
\newblock {H}ate{BERT}: Retraining {BERT} for abusive language detection in {E}nglish.
\newblock In \emph{Proceedings of the 5th Workshop on Online Abuse and Harms (WOAH 2021)}, pages 17--25, Online. Association for Computational Linguistics.

\bibitem[{Chi et~al.(2024)Chi, Karn, Zhan, Smith, Rando, Zhang, Plawiak, Coudert, Upasani, and Pasupuleti}]{chi2024llama}
Jianfeng Chi, Ujjwal Karn, Hongyuan Zhan, Eric Smith, Javier Rando, Yiming Zhang, Kate Plawiak, Zacharie~Delpierre Coudert, Kartikeya Upasani, and Mahesh Pasupuleti. 2024.
\newblock Llama guard 3 vision: Safeguarding human-ai image understanding conversations.
\newblock \emph{arXiv preprint arXiv:2411.10414}.

\bibitem[{Hanu and {Unitary team}(2020)}]{Detoxify}
Laura Hanu and {Unitary team}. 2020.
\newblock Detoxify.
\newblock Github. https://github.com/unitaryai/detoxify.

\bibitem[{Inan et~al.(2023)Inan, Upasani, Chi, Rungta, Iyer, Mao, Tontchev, Hu, Fuller, Testuggine et~al.}]{inan2023llama}
Hakan Inan, Kartikeya Upasani, Jianfeng Chi, Rashi Rungta, Krithika Iyer, Yuning Mao, Michael Tontchev, Qing Hu, Brian Fuller, Davide Testuggine, and 1 others. 2023.
\newblock Llama guard: Llm-based input-output safeguard for human-ai conversations.
\newblock \emph{arXiv preprint arXiv:2312.06674}.

\bibitem[{Lin et~al.(2023)Lin, Wang, Tong, Wang, Guo, Wang, and Shang}]{lin2023toxicchat}
Zi~Lin, Zihan Wang, Yongqi Tong, Yangkun Wang, Yuxin Guo, Yujia Wang, and Jingbo Shang. 2023.
\newblock Toxicchat: Unveiling hidden challenges of toxicity detection in real-world user-ai conversation.
\newblock \emph{arXiv preprint arXiv:2310.17389}.

\bibitem[{Llama~Team(2024)}]{dubey2024llama3herdmodels}
AI~@~Meta Llama~Team. 2024.
\newblock \href {https://arxiv.org/abs/2407.21783} {The llama 3 herd of models}.
\newblock \emph{Preprint}, arXiv:2407.21783.

\bibitem[{Microsoft(2024)}]{azure_ai_content_safety_2024}
Microsoft. 2024.
\newblock \href {https://azure.microsoft.com/en-us/products/ai-services/ai-content-safety} {Azure ai content safety: Detect and moderate harmful content in text and images}.
\newblock Accessed: 2025-01-08.

\bibitem[{OpenAI(2024)}]{openai_moderation_2024}
OpenAI. 2024.
\newblock \href {https://platform.openai.com/docs/guides/moderation/} {Moderation api: A tool for content moderation in language models}.
\newblock Accessed: 2025-01-08.

\bibitem[{Perspective(2024)}]{perspective_api_2024}
Perspective. 2024.
\newblock \href {https://perspectiveapi.com/} {Perspective api: A tool for toxicity detection in online content}.
\newblock Accessed: 2025-01-08.

\bibitem[{Rebedea et~al.(2023)Rebedea, Dinu, Sreedhar, Parisien, and Cohen}]{rebedea2023nemo}
Traian Rebedea, Razvan Dinu, Makesh Sreedhar, Christopher Parisien, and Jonathan Cohen. 2023.
\newblock Nemo guardrails: A toolkit for controllable and safe llm applications with programmable rails.
\newblock \emph{arXiv preprint arXiv:2310.10501}.

\bibitem[{R{\"o}ttger et~al.(2023)R{\"o}ttger, Kirk, Vidgen, Attanasio, Bianchi, and Hovy}]{rottger2023xstest}
Paul R{\"o}ttger, Hannah~Rose Kirk, Bertie Vidgen, Giuseppe Attanasio, Federico Bianchi, and Dirk Hovy. 2023.
\newblock Xstest: A test suite for identifying exaggerated safety behaviours in large language models.
\newblock \emph{arXiv preprint arXiv:2308.01263}.

\bibitem[{Selvaraju et~al.(2017)Selvaraju, Cogswell, Das, Vedantam, Parikh, and Batra}]{selvaraju2017grad}
Ramprasaath~R Selvaraju, Michael Cogswell, Abhishek Das, Ramakrishna Vedantam, Devi Parikh, and Dhruv Batra. 2017.
\newblock Grad-cam: Visual explanations from deep networks via gradient-based localization.
\newblock In \emph{Proceedings of the IEEE international conference on computer vision}, pages 618--626.

\bibitem[{Shrikumar et~al.(2016)Shrikumar, Greenside, Shcherbina, and Kundaje}]{shrikumar2016not}
Avanti Shrikumar, Peyton Greenside, Anna Shcherbina, and Anshul Kundaje. 2016.
\newblock Not just a black box: Learning important features through propagating activation differences.
\newblock \emph{arXiv preprint arXiv:1605.01713}.

\bibitem[{Sundararajan et~al.(2017)Sundararajan, Taly, and Yan}]{pmlr-v70-sundararajan17a}
Mukund Sundararajan, Ankur Taly, and Qiqi Yan. 2017.
\newblock \href {https://proceedings.mlr.press/v70/sundararajan17a.html} {Axiomatic attribution for deep networks}.
\newblock In \emph{Proceedings of the 34th International Conference on Machine Learning}, volume~70 of \emph{Proceedings of Machine Learning Research}, pages 3319--3328. PMLR.

\bibitem[{Team et~al.(2024)Team, Mesnard, Hardin, Dadashi, Bhupatiraju, Pathak, Sifre, Rivi{\`e}re, Kale, Love et~al.}]{team2024gemma}
Gemma Team, Thomas Mesnard, Cassidy Hardin, Robert Dadashi, Surya Bhupatiraju, Shreya Pathak, Laurent Sifre, Morgane Rivi{\`e}re, Mihir~Sanjay Kale, Juliette Love, and 1 others. 2024.
\newblock Gemma: Open models based on gemini research and technology.
\newblock \emph{arXiv preprint arXiv:2403.08295}.

\bibitem[{Touvron et~al.(2023)Touvron, Martin, Stone, Albert, Almahairi, Babaei, Bashlykov, Batra, Bhargava, Bhosale et~al.}]{touvron2023llama}
Hugo Touvron, Louis Martin, Kevin Stone, Peter Albert, Amjad Almahairi, Yasmine Babaei, Nikolay Bashlykov, Soumya Batra, Prajjwal Bhargava, Shruti Bhosale, and 1 others. 2023.
\newblock Llama 2: Open foundation and fine-tuned chat models.
\newblock \emph{arXiv preprint arXiv:2307.09288}.

\bibitem[{Wang et~al.(2024)Wang, Haddow, Birch, and Peng}]{wang-etal-2024-assessing}
Weixuan Wang, Barry Haddow, Alexandra Birch, and Wei Peng. 2024.
\newblock \href {https://doi.org/10.18653/v1/2024.naacl-long.46} {Assessing factual reliability of large language model knowledge}.
\newblock In \emph{Proceedings of the 2024 Conference of the North American Chapter of the Association for Computational Linguistics: Human Language Technologies (Volume 1: Long Papers)}, pages 805--819, Mexico City, Mexico. Association for Computational Linguistics.

\bibitem[{Xiang et~al.(2024)Xiang, Zheng, Li, Hong, Li, Xie, Zhang, Xiong, Xie, Yang et~al.}]{xiang2024guardagent}
Zhen Xiang, Linzhi Zheng, Yanjie Li, Junyuan Hong, Qinbin Li, Han Xie, Jiawei Zhang, Zidi Xiong, Chulin Xie, Carl Yang, and 1 others. 2024.
\newblock Guardagent: Safeguard llm agents by a guard agent via knowledge-enabled reasoning.
\newblock \emph{arXiv preprint arXiv:2406.09187}.

\bibitem[{Xie et~al.(2024)Xie, Fang, Pi, and Gong}]{xie2024gradsafe}
Yueqi Xie, Minghong Fang, Renjie Pi, and Neil Gong. 2024.
\newblock Gradsafe: Detecting unsafe prompts for llms via safety-critical gradient analysis.
\newblock \emph{arXiv preprint arXiv:2402.13494}.

\bibitem[{Yang et~al.(2024)Yang, Yang, Zhang, Hui, Zheng, Yu, Li, Liu, Huang, Wei et~al.}]{yang2024qwen2}
An~Yang, Baosong Yang, Beichen Zhang, Binyuan Hui, Bo~Zheng, Bowen Yu, Chengyuan Li, Dayiheng Liu, Fei Huang, Haoran Wei, and 1 others. 2024.
\newblock Qwen2. 5 technical report.
\newblock \emph{arXiv preprint arXiv:2412.15115}.

\bibitem[{Zhang et~al.(2024)Zhang, Lu, Ma, Zhang, Li, Ke, Sun, Sha, Sui, Wang, and Huang}]{DBLP:conf/emnlp/ZhangLMZLKSSSWH24}
Zhexin Zhang, Yida Lu, Jingyuan Ma, Di~Zhang, Rui Li, Pei Ke, Hao Sun, Lei Sha, Zhifang Sui, Hongning Wang, and Minlie Huang. 2024.
\newblock Shieldlm: Empowering llms as aligned, customizable and explainable safety detectors.
\newblock In \emph{Findings of the Association for Computational Linguistics: {EMNLP} 2024, Miami, Florida, USA, November 12-16, 2024}, pages 10420--10438. Association for Computational Linguistics.

\end{thebibliography}
\end{document}